\pdfoutput=1
\documentclass[11pt, a4paper, copyright, gdm]{google}

\usepackage[authoryear, sort&compress, round]{natbib}
\expandafter\def\expandafter\UrlBreaks\expandafter{\UrlBreaks\do-}
\usepackage{booktabs}
\usepackage{graphicx}
\usepackage{float}
\usepackage{listings}

\usepackage{xcolor}

\uselogo{} 

\title{MKA: Evaluating Multimodal Knowledge Agents on end-to-end professional tasks}

\correspondingauthor{\string{animasingh, suhasyogin, weicheng, rohandoshi, bencaine\string}@google.com}

\renewcommand{\today}{2026-11-10}

\author{\textbf{Anima Singh}\textsuperscript{\normalfont *}}
\author{\textbf{Suhas Yogin}\textsuperscript{\normalfont *}}
\author{\textbf{Weicheng Kuo}\textsuperscript{\normalfont *}}
\author{\textbf{Ivor Rendulic}}
\author{\textbf{Ivan Jurin}}
\author{\textbf{Antoine Yang}}
\author{\textbf{Oran Kelly}}
\author{\textbf{Rohan Doshi}}
\author{Hammad Ayyubi}
\author{Ioana Bica}
\author{Sergi Caelles}
\author{Matthew Conniff}
\author{Dan Graur}
\author{Anastasija Ili\'c}
\author{Ahmet Iscen}
\author{Chao Jia}
\author{Tomasz K\k{e}pa}
\author{Chunyuan Li}
\author{Noah \'O Donnaile}
\author{Filip Pavetic}
\author{Goran Petrovic}
\author{Khalid Salama}
\author{Zhicheng Wang}
\author{Polina Zablotskaia}
\author{Yiqian Zhang}
\author{Ming Zhao}
\author{\textbf{Benjamin Caine}}
\author{\textbf{Mario Lu\v{c}i\'c}}
\author{Melvin Johnson}
\author{Radu Soricut}
\author{Jean-Baptiste Alayrac}

\affil{Google DeepMind}

\begin{abstract}
We introduce the Multimodal Knowledge Agent (MKA) benchmark, which measures how well autonomous agents perform end-to-end professional knowledge work that depends on visual understanding. 
Each task requires an agent to extract and combine information from multiple visually dense files, such as CAD blueprints, electronics schematics, radiographs, complex charts and geospatial maps, and to produce a standalone deliverable in \texttt{.xlsx}, \texttt{.pptx}, \texttt{.pdf}, \texttt{.docx}, \texttt{.html} or \texttt{.png} format.
Existing benchmarks leave this capability largely untested: removing image-viewing
tools leaves Gemini~4 Carbon's score on GDPval~\citep{patwardhan2025gdpval} unchanged
(Section~\ref{sec:vision-ablation}),
and single-format evaluations such as GDP.pdf~\citep{garre2026gdppdf} and GDP.xlsx~\citep{surgeai2026gdpxlsx}
do not require reasoning across files or modalities. MKA contains 200 tasks across 10
economic sectors.
Deliverables are graded against expert-weighted rubrics by a sandboxed agentic autorater
that agrees with human domain experts on 96.2\% of criterion judgments in our expert-graded study (Cohen's $\kappa = 0.91$), scored by
weighted satisfaction rate (WSR) overall and on frontier criteria unsaturated by
prior-generation models (Climb-WSR).
Across the frontier models we evaluate (Gemini~4 Carbon, Claude Fable~5.2, Claude
Opus~5.5, and GPT-6 Astra), Claude Fable~5.2 (56.6\% Climb-WSR, 73.3\% Overall WSR) and
Gemini~4 Carbon (54.3\% Climb-WSR, 71.9\% Overall WSR) achieve the highest scores, while Carbon uses $5.7\times$ fewer tokens. Even for the strongest models, Climb-WSR remains
below 57\%, leaving substantial headroom for multimodal agents.

\end{abstract}
\makeatletter
\g@addto@macro\ps@firststyle{%
  \fancyfoot[L,C,R]{}%
}
\newcommand{\firstpagefooterblock}{%
  \begingroup
  \NoHyper
  \global\let\orig@footnoterule\footnoterule
  \gdef\footnoterule{\kern-3pt{\color{gray}\hrule width \textwidth height 1pt}\kern 4pt\global\let\footnoterule\orig@footnoterule}%
  \long\def\@makefntext##1{\parindent 0pt\parskip 2pt\noindent\footerfont##1}%
  \footnotetext{%
    \textsuperscript{*}Equal contribution (co-first authors): Anima Singh (research \& evaluation lead), Suhas Yogin (end-to-end dataset \& release lead), and Weicheng Kuo (early data \& task design lead). Core contributors are shown in \textbf{bold}, followed by contributors in alphabetical order and leadership.\par
    {\itshape Corresponding author(s): \the\correspondingauthor}%
    \ifthenelse{\boolean{copyright}}{\\\copyrightext}{}%
  }%
  \endNoHyper
  \endgroup
}
\makeatother

\begin{document}
\maketitle
\enlargethispage{38pt}
\firstpagefooterblock
\section{Introduction}
\label{sec:intro}
\begin{figure}[t]
  \centering
  \includegraphics[width=\linewidth]{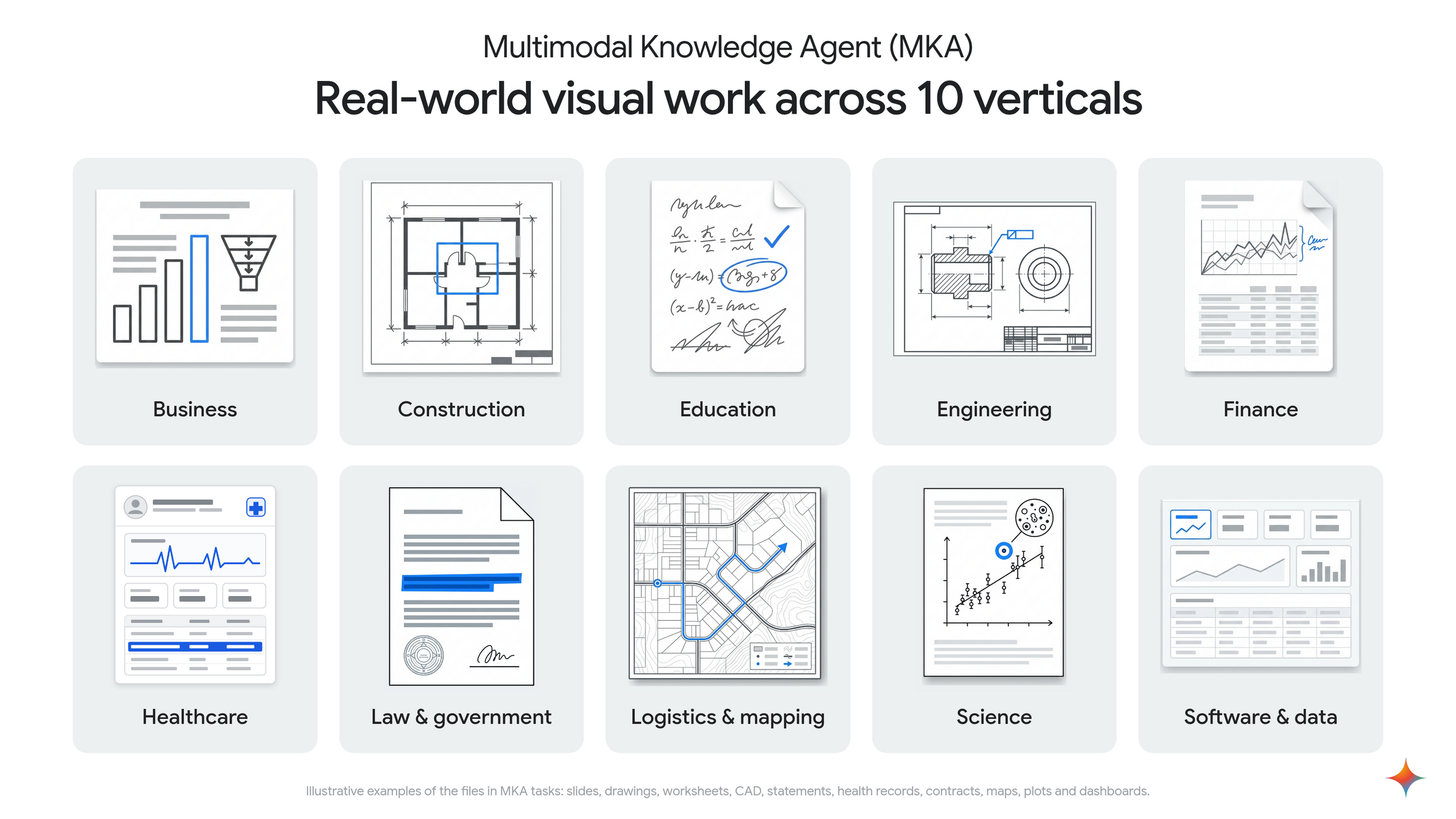}
  \caption{Overview of the 10 economic sectors in MKA (200 tasks total, 20 per sector), spanning blueprints, schematics, financial charts, clinical records, contracts, geospatial maps, scientific plots, and interactive dashboards.}
  \label{fig:domains}
\end{figure}
Real-world knowledge work is deeply visual. Professionals routinely read technical diagrams and schematics, interpret complex charts~\citep{wang2024charxiv}, and work with messy scanned documents~\citep{ma2024mmlongbenchdoc}---all while visually verifying the formatting and layout of their own work products.
Yet today's frontier agents, shaped by coding-centric training~\citep{wang2024codeact,jimenez2024swebench,merrill2026terminalbench}, exhibit a surprising degree of text-and-code bias rather than using vision as a tool. Given a workspace of files, they instinctively convert documents to plain text, Markdown, or CSVs, and often rely heavily on brittle OCR engines or traditional computer-vision scripts despite possessing far stronger native visual capabilities \citep{deepmind2025gemini3,google2026agenticvision,anthropic2026claudeopus46,openai2026gpt5,openai2025o3,meta2026muse,bai2025qwen3vl,zhu2025internvl3,meta2025llama4}---and even when looking at images directly, models remain prone to geometric blindness and strong text-prior bias~\citep{rahmanzadehgervi2024vlmsareblind,vo2025vlmsarebiased}.

Existing agentic benchmarks fail to expose this blind spot (Table~\ref{tab:benchmark_comparison}; Section~\ref{sec:related_work}). Multi-file workplace evaluations such as GDPval~\citep{patwardhan2025gdpval} and APEX-Agents~\citep{vidgen2026apexagents}, either allow task-critical signal to be extracted via scripts with negligible score loss, or focus on text- and spreadsheet-heavy domains. Meanwhile, visual QA suites \citep{yue2024mmmupro,wang2024charxiv,chen2024mmstar,fu2024blink,garre2026gdppdf} test single-turn text responses without tool use, spreadsheet benchmarks \citep{surgeai2026gdpxlsx,ma2024spreadsheetbench,zhu2026spreadsheetbench2} remain strictly tabular, browsing benchmarks~\citep{wei2025browsecomp,li2025mmbrowsecomp} output short exact-match strings, and GUI benchmarks~\citep{xie2024osworld,li2025screenspotpro} entangle reasoning with pixel-level clicking.

To close this gap, we introduce \textbf{MKA (Multimodal Knowledge Agent)}, an end-to-end benchmark of 200 professional workflows across 10 evenly balanced sectors and 83 occupations (Figure~\ref{fig:domains}; Section~\ref{sec:dataset}), alongside 10 disjoint open showcase tasks (Section~\ref{sec:release}). Each task provides 2 to 45 heterogeneous files (mean 7.3) with visual artifacts---such as scaled floor plans, electronics schematics, route maps, or radiographs---where flattening files to text destroys crucial information. Standalone file deliverables (\texttt{.xlsx}, \texttt{.pptx}, \texttt{.pdf}, \texttt{.docx}, \texttt{.html}, \texttt{.png}, \texttt{.txt}) are required in 88\% of tasks, making vision essential both to read inputs and to create final outputs.

Controlled tool ablations in our standardized Stirrup harness~\citep{artificialanalysis2025stirrup} show that vision in MKA is load-bearing. Removing image-viewing tools (\texttt{view\_image}) from Gemini~4 Carbon leaves GDPval quality and compute unchanged, whereas on MKA it reduces Climb-WSR by $\sim$5.9 percentage points and inflates agent turns by 1.28$\times$ and output tokens by 1.40$\times$ as the model falls back on coding workarounds. These workarounds recover part of the score on clean drawings but fail outright on photographs and medical images  (Section~\ref{sec:vision-ablation}).

To grade binary deliverables at scale without repeating the text-extraction mistake, we pair Stirrup with a sandboxed Agentic Autorater~\citep{zhuge2025agentasajudge,chen2024mllmasajudge} that visually renders candidate files and programmatically inspects formulas and document internals, reaching 96.2\% pairwise criterion agreement with vetted human experts in a 50-task study ($\kappa = 0.91$; task-level $r = 0.966$; Section~\ref{sec:human-correlation}).
Evaluating leading frontier models (Gemini~4 Carbon, Claude Fable~5.2, Claude Opus~5.5, and GPT-6 Astra) under identical harness budgets, Claude Fable~5.2 (56.6\% Climb-WSR, 73.3\% Overall WSR) and Gemini~4 Carbon (54.3\% Climb-WSR, 71.9\% Overall WSR) achieve the strongest deliverable quality within statistical noise, with Carbon using $5.7\times$ fewer tokens and $6.3\times$ lower API cost (Section~\ref{sec:results}), while top Climb-WSR near 57\% highlights substantial remaining headroom for genuinely multimodal agents.

\begin{table}[t]
\centering
\caption{Comparison of MKA against representative frontier benchmarks across multimodal, browsing, GUI, tabular, and professional knowledge-work evaluations.}
\label{tab:benchmark_comparison}
\footnotesize
\setlength{\tabcolsep}{1.5pt}
\renewcommand{\arraystretch}{1.2}
\begin{tabular*}{\textwidth}{@{\extracolsep{\fill}}p{0.31\textwidth}ccccc@{}}
\toprule
 & \textbf{Load-Bearing} & \textbf{Multi-Step} & \textbf{Heterogeneous} & \textbf{Open-Ended} & \textbf{Visual + Code} \\
\textbf{Benchmark} & \textbf{Vision} & \textbf{Tool Use} & \textbf{Multi-File} & \textbf{Deliverables} & \textbf{Grading} \\
\midrule
\raggedright MMMU-Pro / CharXiv~\citep{yue2024mmmupro,wang2024charxiv} & Yes & No & No & No & No \\
\raggedright GDP.pdf~\citep{garre2026gdppdf} & Yes & No & No (PDFs Only) & No & No \\
\raggedright GDP.xlsx~\citep{surgeai2026gdpxlsx} & No & Yes & No (Single File) & No & No \\
\raggedright BrowseComp / MM-BrowseComp~\citep{wei2025browsecomp,li2025mmbrowsecomp} & Partial & Yes & No & No & No \\
\raggedright OSWorld~\citep{xie2024osworld} & Yes (GUI) & Yes & Partial & Partial & No \\
\raggedright SpreadsheetBench~2~\citep{zhu2026spreadsheetbench2} & No (Inputs) & Yes & No (Tabular) & Yes (\texttt{.xlsx}) & Partial \\
\raggedright APEX-Agents~\citep{vidgen2026apexagents} & Partial & Yes & Yes & Partial & No \\
\raggedright GDPval~\citep{patwardhan2025gdpval} & No (Bypassable) & Yes & Yes & Yes & No \\
\midrule
\textbf{MKA (Ours)} & \textbf{Yes} & \textbf{Yes} & \textbf{Yes} & \textbf{Yes (88\%)} & \textbf{Yes} \\
\bottomrule
\end{tabular*}
\end{table}

\section{The MKA Dataset}
\label{sec:dataset}

MKA consists of 200 real-world tasks written and verified by vetted domain
professionals. The tasks are split evenly across 10 sectors (20 tasks each) and
cover 83 occupations, including financial analysts, licensed architects, electrical
and mechanical engineers, GIS surveyors, data scientists, attorneys and physicians.
Each author has at least five years of professional experience, and every task was
reviewed for realism and accuracy by a second expert with the same job profile.

Each task has a single prompt that specifies the user scenario, the requirements
and the target deliverable. An expert spent 3 to 8 hours on each task: assembling a
realistic multi-file input bundle, solving the task to produce golden reference
deliverables, and writing a grading rubric with a golden justification for every
criterion.

\begin{figure}[t]
  \centering
  \includegraphics[width=0.92\linewidth]{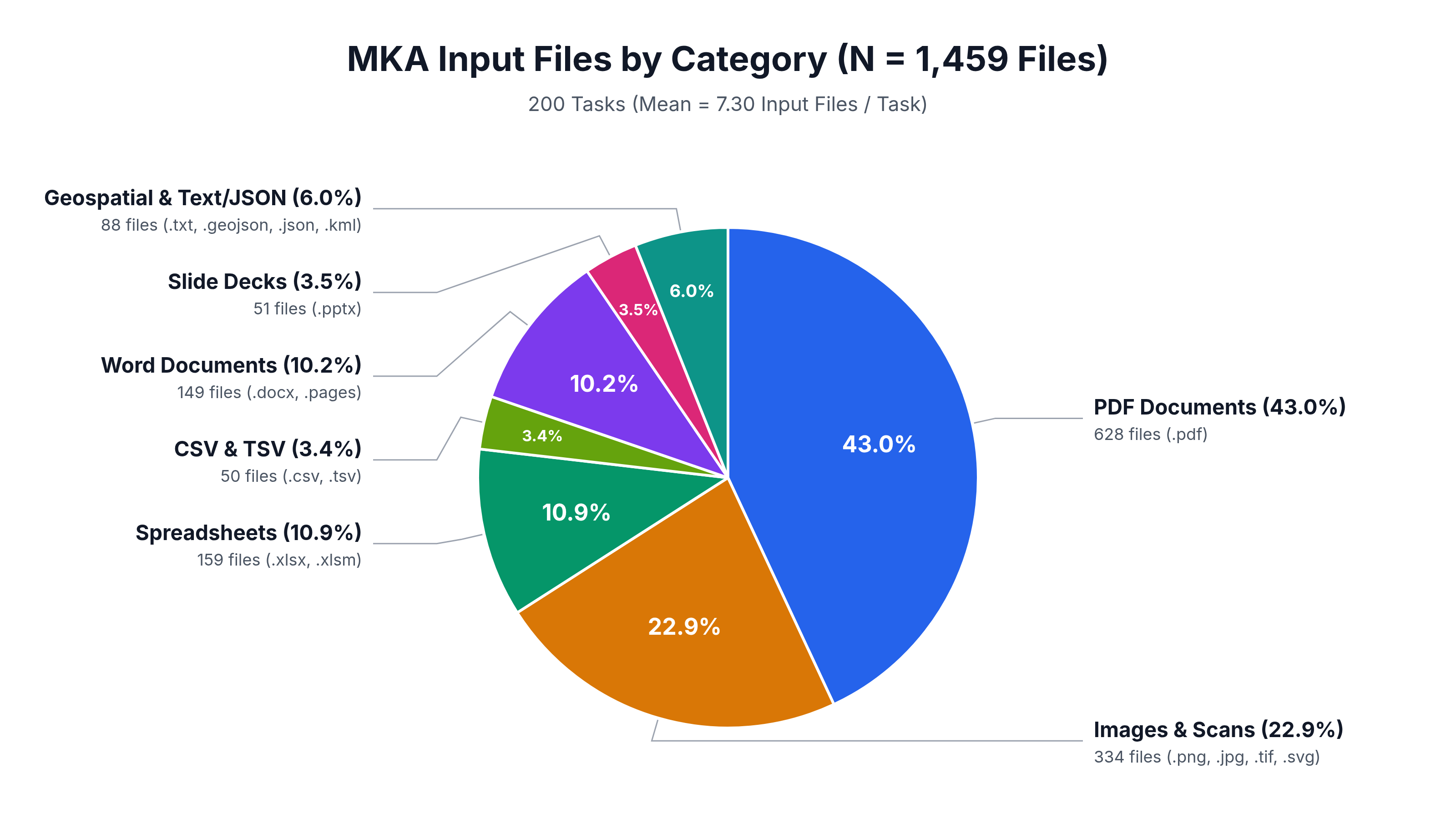}
  \caption{Distribution of the 1,459 input files across the 200 MKA tasks by format category.}
  \label{fig:input_files}
\end{figure}

\paragraph{Input files.}
Each task includes 2 to 45 input files (1,459 in total; mean 7.30, median 6.0), and
solving it requires combining information across visual, tabular, document and
geospatial formats. Figure~\ref{fig:input_files} breaks down the 1,459 input files by
format category (percentages are shares of total input files):
\begin{itemize}
  \item \textbf{PDF documents (43.0\%, 628 files):} architectural elevations and floor
        plans, CAD and piping schematics, SEC Form 10-K filings, commercial leases,
        clinical operative reports and regulatory manuals (\texttt{.pdf}).
  \item \textbf{Images and scans (22.9\%, 334 files):} floor plans, circuit and
        mechanical drawings, CT scans and radiographs, GIS route maps, digital
        elevation models and scanned handwritten forms (217 \texttt{.png}, 105
        \texttt{.jpg}/\texttt{.jpeg}, 6 \texttt{.tif}, 3 \texttt{.gif}, 3
        \texttt{.svg}).
  \item \textbf{Spreadsheets (10.9\%, 159 files):} multi-tab Excel workbooks
        containing balance sheets and operating models, claims and exposure registers,
        engineering bills of materials and project Gantt charts (158 \texttt{.xlsx},
        1 \texttt{.xlsm}).
  \item \textbf{Word documents (10.2\%, 149 files):} standard operating procedures,
        purchase agreements, clinical coding crosswalks, witness statements and report
        templates (148 \texttt{.docx}, 1 \texttt{.pages}).
  \item \textbf{Geospatial and text/JSON (6.0\%, 88 files):} municipal boundary and
        route layers, LiDAR and cartography specifications, EEG metadata, email
        threads and correspondence logs (42 \texttt{.txt}, 21 \texttt{.geojson}, 10
        \texttt{.json}, 3 \texttt{.md}, 3 \texttt{.eml}, 2 \texttt{.html}, 2
        \texttt{.kml} and 5 shapefile/world-file/script assets).
  \item \textbf{Slide decks (3.5\%, 51 files):} executive board briefings, investment
        mandates, network architecture boards and presentation templates
        (\texttt{.pptx}).
  \item \textbf{CSV and TSV (3.4\%, 50 files):} delimited telemetry logs, treasury
        rate series, fleet dispatch manifests and clinical trial tables (47
        \texttt{.csv}, 3 \texttt{.tsv}).
\end{itemize}

\begin{figure}[t]
  \centering
  \includegraphics[width=0.92\linewidth]{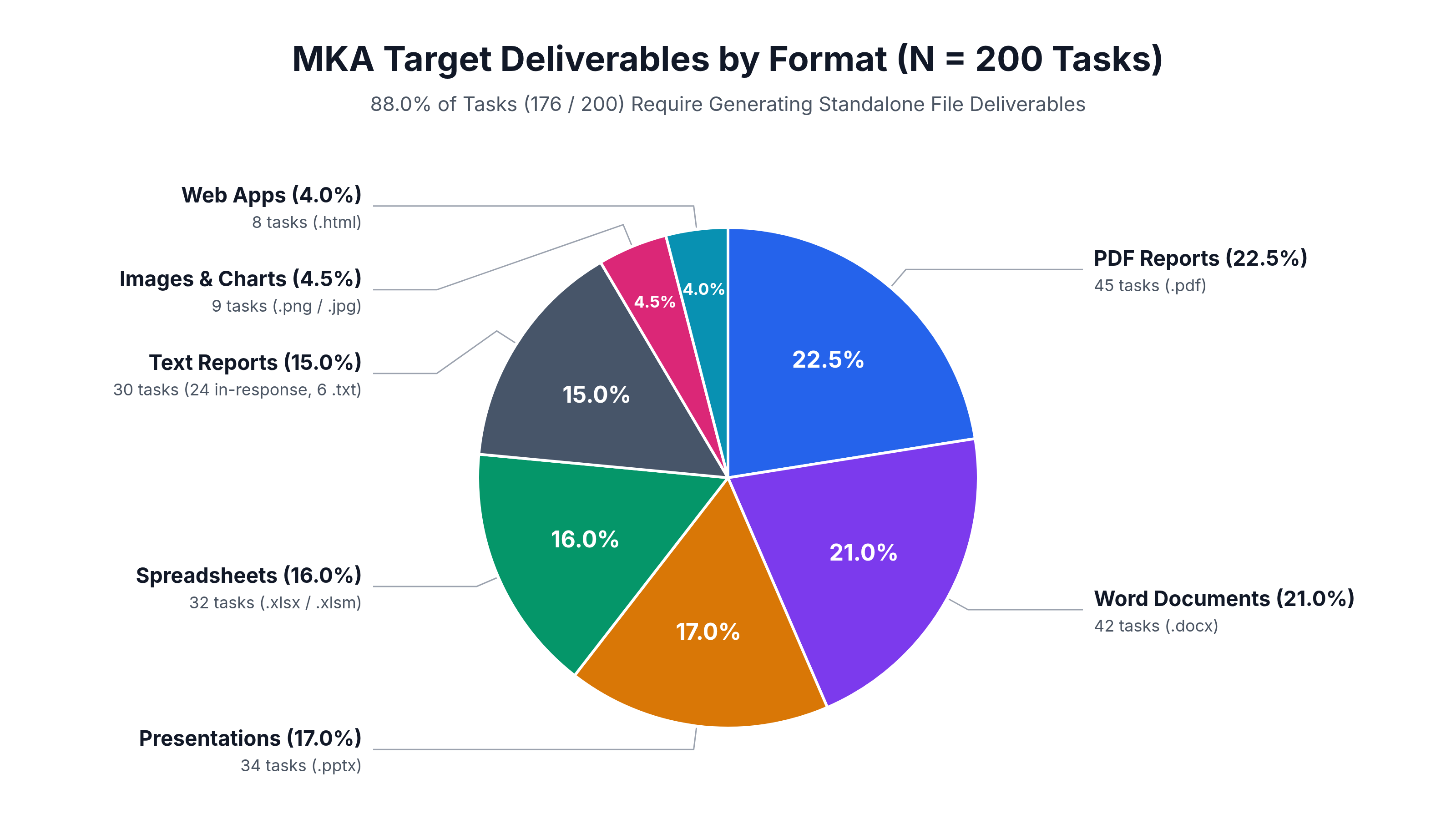}
  \caption{Distribution of primary deliverable formats across the 200 MKA tasks.}
  \label{fig:deliverables}
\end{figure}

\paragraph{Deliverables.}
Of the 200 tasks, 176 (88.0\%) require the agent to create or modify a standalone
file, while the remaining 24 (12.0\%) are answered directly in the text response.
Figure~\ref{fig:deliverables} shows the distribution of primary deliverables across
all 200 tasks (percentages sum to 100.0\%):
\begin{itemize}
  \item \textbf{PDF reports (22.5\%, 45 tasks):} formatted engineering evaluations,
        property appraisals, clinical SOPs and audit memoranda (\texttt{.pdf}).
  \item \textbf{Word documents (21.0\%, 42 tasks):} structured technical reports,
        legal and regulatory reviews, and feasibility studies (\texttt{.docx}).
  \item \textbf{Presentations (17.0\%, 34 tasks):} multi-slide executive briefings,
        investment proposals and training decks (\texttt{.pptx}).
  \item \textbf{Spreadsheets (16.0\%, 32 tasks):} multi-tab financial projections,
        tax and accounting schedules, and engineering workbooks (\texttt{.xlsx},
        \texttt{.xlsm}).
  \item \textbf{Text reports (15.0\%, 30 tasks):} structured written analyses returned
        either directly in the text response (24 tasks, 12.0\%) or as standalone
        plain-text files (6 tasks, 3.0\%, \texttt{.txt}).
  \item \textbf{Images and charts (4.5\%, 9 tasks):} high-resolution maps, diagrams
        and visual figures (7 \texttt{.png}, 2 \texttt{.jpg}).
  \item \textbf{Web apps (4.0\%, 8 tasks):} self-contained interactive HTML dashboards
        and visual applications (\texttt{.html}).
\end{itemize}

\paragraph{Rubrics.}
Across the 200 tasks, MKA contains 14,284 rubric criteria (mean 71.4 per task, median
68.5, range 20--150):
\begin{itemize}
  \item \textbf{Target:} 91.3\% of criteria (13,037) evaluate the 176 tasks requiring
        standalone file deliverables, while 8.7\% (1,247) evaluate the 24 text-response
        tasks (1,735 criteria, or 12.1\%, when including the 6 standalone \texttt{.txt}
        tasks).
  \item \textbf{Weights and priority:} each criterion $j$ of task $i$ has an integer
        weight $w_{ij} \in \{1,\dots,5\}$ (21.6\%, 26.2\%, 26.1\%, 21.4\% and 4.7\%
        for weights 1--5, respectively) used to compute the Weighted Satisfaction Rate
        (Section~\ref{sec:metrics}), and is labeled as a primary objective (94.5\%,
        13,497 criteria) or a secondary objective (5.5\%, 787 criteria).
  \item \textbf{Dependencies:} 29.5\% of criteria (4,217 across 190 tasks) list
        explicit prerequisite criteria in their metadata, forming a directed acyclic
        graph (DAG) within each task that records how downstream conclusions depend on
        upstream visual extractions and calculations (during grading, the autorater
        evaluates each criterion independently and WSR aggregates all criteria by
        $w_{ij}$).
  \item \textbf{Golden justifications:} 99.6\% of criteria (14,229 of 14,284; 100\%
        across 199 of the 200 tasks) include an expert-written golden justification
        that cites the source location of the ground truth and, where applicable, its
        derivation.
\end{itemize}

\paragraph{Sectors.}
Figure~\ref{fig:domains} illustrates how the 200 tasks are distributed evenly across
10 sectors (20 tasks, or 10.0\%, per sector) and 83 occupations. The bullets below
highlight a few reference job profiles in each sector:
\begin{itemize}
  \item \textbf{Business:} management consultants, marketing and
        media analysts, customer support specialists, revenue management analysts and
        procurement specialists.
  \item \textbf{Construction:} licensed architects, construction
        estimators, quantity surveyors, real estate appraisers and civil or fire
        protection engineers.
  \item \textbf{Education:} K--12 and university instructors,
        curriculum compliance analysts, academic moderators, school psychologists and
        training specialists.
  \item \textbf{Engineering:} mechanical and CAD engineers,
        electrical and RF engineers, chemical process engineers, power systems
        engineers and manufacturing quality engineers.
  \item \textbf{Finance:} financial and investment analysts,
        accountants and auditors, property and marine underwriters, credit analysts
        and tax preparers.
  \item \textbf{Healthcare:} diagnostic and pediatric radiologists,
        radiation oncologists, telestroke neurologists, hospital administrators and
        medical coding or claims specialists.
  \item \textbf{Law and government:} immigration, corporate, real
        estate and patent attorneys, regulatory analysts, municipal procurement
        officers and government administrators.
  \item \textbf{Logistics and mapping:} GIS analysts, land
        surveyors, urban and regional transport planners, precision agriculture
        analysts and logistics analysts.
  \item \textbf{Science:} clinical trial researchers, medicinal and
        analytical chemists, molecular and wildlife biologists, materials scientists
        and EEG analysts.
  \item \textbf{Software and data:} front-end engineers, data
        analysts, data engineers, software QA engineers and radio or network
        engineers.
\end{itemize}

\begin{figure}[t]
  \centering
  \includegraphics[width=\textwidth]{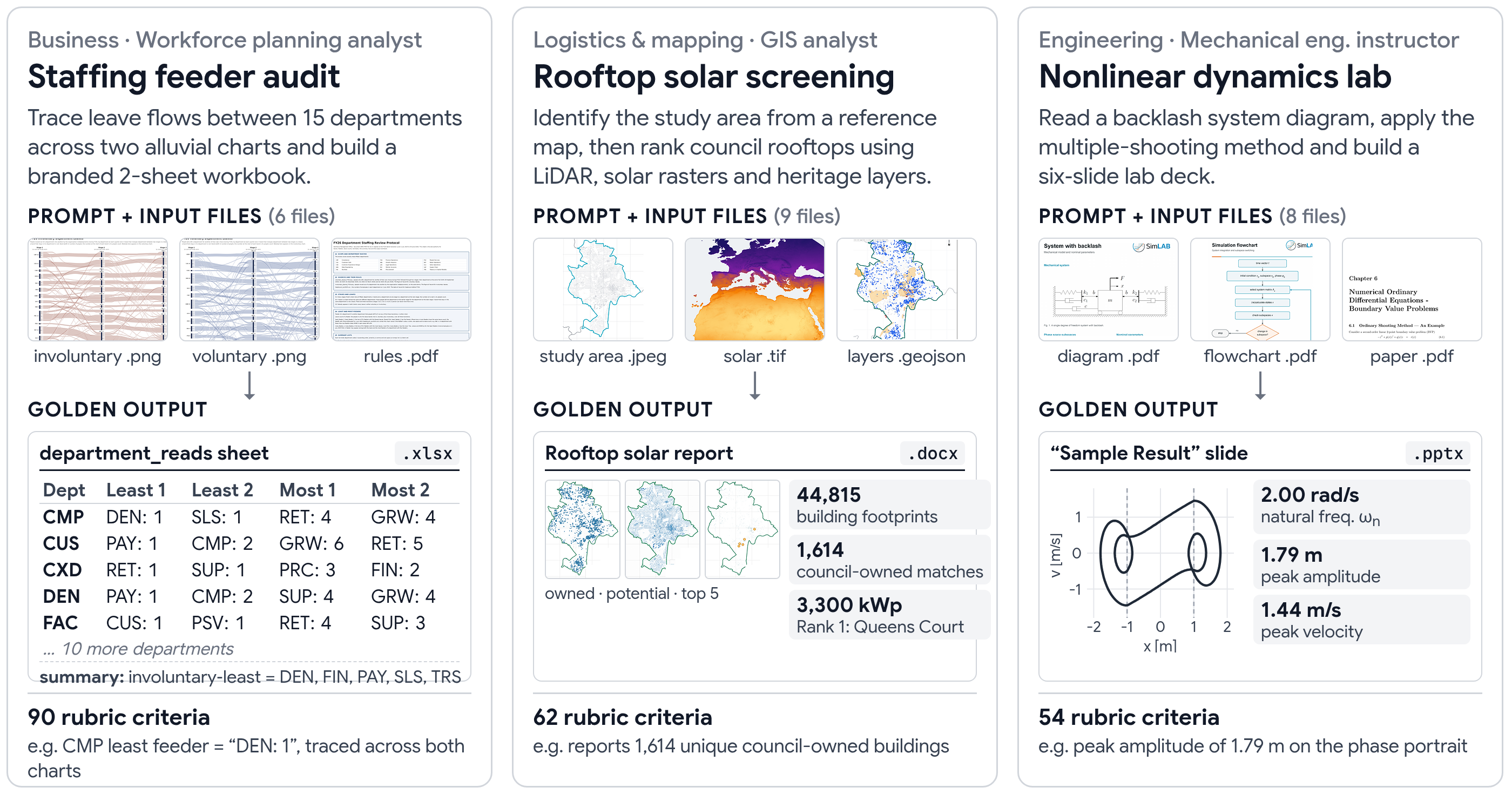}
  \caption{Three example tasks from the open MKA-Showcase split. Each card shows the
  sector and expert role, a summary of the prompt, three of the task's input files, a
  preview of the human golden deliverable, and the number of rubric criteria with one
  example criterion: a department staffing feeder audit (workforce planning analyst), a
  council rooftop solar screening (GIS analyst), and a nonlinear dynamics lab deck
  (mechanical engineering instructor).}
  \label{fig:examples}
\end{figure}

\paragraph{Reference examples.}
Figure~\ref{fig:examples} shows three example tasks from the open MKA-Showcase split (Section~\ref{sec:release}):
\begin{itemize}
  \item \textbf{Workforce planning analyst (Business).} The agent conducts a fiscal-year
        department staffing review across 6 input files: quarterly voluntary and
        involuntary leave alluvial charts (\texttt{.png}), baseline headcount records
        (\texttt{.csv}), staffing and tie-break rules (\texttt{.pdf}), a corporate
        branding template (\texttt{.xlsx}) and a company logo (\texttt{.jpg}). It must
        trace leave flows between 15 departments across both charts and produce a
        branded 2-sheet workbook (\texttt{.xlsx}) ranking each department's least and
        most frequent feeders (90 rubric criteria).
  \item \textbf{GIS analyst (Logistics and mapping).} The agent prepares a municipal
        rooftop solar assessment across 9 geospatial and tabular files. It identifies
        the study area (City of Nottingham) from an unlabeled reference map
        (\texttt{.jpeg}), joins administrative, building-footprint and heritage layers
        (\texttt{.geojson}) against 1\,m LiDAR elevation and solar-irradiance rasters
        (\texttt{.tif}), a council asset register (\texttt{.xlsx}) and screening rules
        (\texttt{.txt}), and produces an interleaved report (\texttt{.docx}) with maps
        and a ranked table of the top five council buildings (62 rubric criteria).
  \item \textbf{Mechanical engineering instructor (Engineering).} The agent builds a
        nonlinear dynamics lab deck across 8 input files. It reads bilateral backlash
        geometry and parameters from a mechanical schematic (\texttt{.pdf}), follows a
        subspace-switching flowchart and multiple-shooting reference paper
        (\texttt{.pdf}), applies presentation specifications (\texttt{.docx},
        \texttt{.pdf}) and a slide template (\texttt{.pptx}), and solves the periodic
        boundary-value problem to produce a 6-slide presentation (\texttt{.pptx}) with
        phase-portrait charts (54 rubric criteria).
\end{itemize}

\Needspace{7\baselineskip}
\subsection{Public Leaderboard}
\label{sec:release}
To support independent evaluation while preventing benchmark contamination, MKA is accompanied by two disjoint task sets. \textbf{MKA (200 tasks)} is the private evaluation suite reported in this paper, hosted and graded independently on a public leaderboard by Artificial Analysis using the open-source Stirrup framework (Section~\ref{sec:harness}). \textbf{MKA-Showcase (10 tasks)} is a separate, publicly downloadable release of one task per sector with full prompts, input files, and grading rubrics.

\section{Evaluation Setup}
\label{sec:evaluation}

\subsection{Agentic Harness}
\label{sec:harness}

We evaluate every model in Stirrup (\texttt{v0.2.1}) \citep{artificialanalysis2025stirrup}, an open-source,
lightweight agent framework maintained by Artificial Analysis. The harness gives the
agent the task prompt and the paths of the input files as text, so the agent must open
and inspect each file and build the final deliverables through tool calls.
All models use the same harness, system instructions, tool set, container image and
resource budgets; only the model endpoint changes. An externally maintained, minimal
harness avoids model-specific scaffolding, so deliverable quality and token efficiency
are directly comparable across model families.

\paragraph{Agent loop.}
Stirrup runs without human intervention until the agent calls \texttt{finish} or
reaches the limit of 250 turns, with up to a 1M-token context window if available, and an 8-hour wall-clock limit per task. At 70\% of the model's context length, compaction is triggered. If the model replies
without a tool call, the harness prompts it to continue. In the final 20 turns, the
harness appends a warning with the number of remaining turns so the agent can finish
and submit its deliverables.

\paragraph{Sandbox.}
Each task runs in a fresh, isolated Linux container (4~vCPUs; 128~MB base RAM dynamically bursting up to 16~GB) based on a Debian container with Python 3.13 (\texttt{python:3.13-\allowbreak slim-\allowbreak trixie}), with working directory \texttt{/workspace} and input
files staged under \texttt{/workspace/input}. The sandbox has internet access. The image includes common
document, PDF, OCR and rendering tools, including LibreOffice, Tesseract,
\texttt{pymupdf}, \texttt{weasyprint} and Python Office libraries, so agents can parse
inputs and build deliverables without installing software.

\paragraph{Tools.}
Stirrup exposes four tools: \texttt{code\_exec}, \texttt{view\_image},
\texttt{web\_search} and \texttt{finish}. \texttt{code\_exec} runs bash commands and
Python scripts (300\,s per-command timeout), and \texttt{web\_search} issues web queries (180\,s timeout). \texttt{view\_image} loads a
raster image (\texttt{.png}, \texttt{.jpg}, \texttt{.webp}, \texttt{.gif}, \texttt{.bmp}
or \texttt{.tiff}) into the model's visual context. To inspect a document or slide deck
visually, the agent renders pages to \texttt{.png} with \texttt{code\_exec} using command-line tools such as LibreOffice and \texttt{pdftoppm}, and views the pages it needs with \texttt{view\_image}. \texttt{view\_image} is designed to respect the native image resolution of the output file. At the end of a trajectory, \texttt{finish} checks that every declared deliverable exists and ends the session.

\subsection{Agentic Autorater}
\label{sec:autorater}

Most MKA tasks (88.0\%; 91.3\% of rubric criteria) require standalone files such as
spreadsheets (\texttt{.xlsx}), slide decks (\texttt{.pptx}), documents (\texttt{.docx}),
PDFs or images. MKA rubrics check both visual layout and underlying file internals such as spreadsheet formulas. Static text extraction omits charts and formatting, while static page renders hide cell formulas and can clip wide sheets. To give the grader direct access to both, we use an agentic, multimodal judge that can inspect rendered pages and run code to query file contents.

\paragraph{Grading harness.}
The Agentic Autorater is also a Stirrup agent. It runs in its own fresh container with
\texttt{code\_exec}, \texttt{view\_image} and a structured \texttt{submit\_verdicts} finish tool, and without web access.
Its \texttt{/workspace} contains only the candidate's deliverable files and final text
response. The autorater does not receive the task's input files: MKA rubrics are self-contained, with each criterion stating the expected value, tolerance or property directly, so grading checks the deliverable against the rubric without re-solving the task. In a single session, the autorater checks these outputs against the verbatim task prompt and rubric and returns, for each criterion, a pass/fail verdict, a justification and the files it cited (prompt in Appendix~\ref{sec:autorater-prompt}).
We use OpenAI GPT-6 Luna as the autorater model. Using a judge from outside the Google model family avoids favoring Gemini; we also see no sign of same-family preference, as GPT-6 Astra ranks last among the evaluated models.

\subsection{Agreement Between the Autorater and Human Raters}
\label{sec:human-correlation}

We compare the Agentic Autorater's verdicts with independent human ratings using the
pairwise agreement metric of GDPval~\citep[App.~A.6.2]{patwardhan2025gdpval}. For binary pass/fail verdicts, this metric measures the fraction of rater pairs that assign the same verdict on a criterion, averaged across criteria, and compares human--autorater agreement against the inter-human agreement ceiling (Eq.~\ref{eq:agreement}). Matching Eq.~\eqref{eq:task-wsr}, we weight criteria by $w_{ij}$ within each task ($\sum_j w_{ij} a_{ij} / \sum_j w_{ij}$) and average equally across the tasks to obtain $A^{\mathrm{HA}}$ and the inter-human ceiling $A^{\mathrm{HH}}$, with 95\% CIs from resampling the tasks with replacement.

\paragraph{Human study.}
Across 50 MKA tasks spanning 9 verticals, we pair each task with one deliverable produced by the
frontier models. Three domain experts, blind to model identity, grade each deliverable independently: for every
rubric criterion, they inspect the task inputs and the deliverable and record a
pass/fail verdict with a written rationale. Unlike the autorater, human raters also see
the task inputs, so agreement between the two also tests whether the rubrics are sufficient for grading on their own. To focus expert annotation on substantive factual, quantitative, and structural requirements, we filter to high-weight rubric criteria (weights 3--5) and trim redundant passing checks on high-scoring tasks to preserve a balanced pass/fail mix. We also run the autorater three times per deliverable.

\paragraph{Agreement metrics.}
For each criterion $c$, we compute human--autorater agreement as
\begin{equation}
  A_c^{\mathrm{HA}} = \mathbb{E}_{i,j}\bigl[1 - \lvert H_{c,i} - R_{c,j} \rvert\bigr],
  \label{eq:agreement}
\end{equation}
where $H_{c,i}, R_{c,j} \in \{0, 1\}$ are the verdicts of human rater $i$ and autorater
run $j$ on criterion $c$. Weighting $A_c^{\mathrm{HA}}$ by $w_{ij}$ within each task and
averaging equally across tasks yields $A^{\mathrm{HA}}$. Analogously, we compute the human ceiling $A^{\mathrm{HH}}$ over distinct
pairs of human raters ($i \neq i'$), with 95\% confidence intervals from a
bootstrap over tasks.

\begin{figure}[t]
  \centering
  \includegraphics[width=0.55\textwidth]{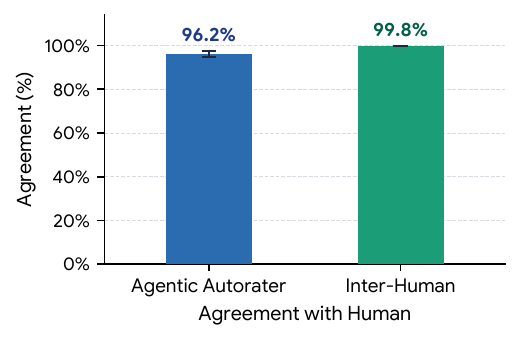}
  \caption{Pairwise criterion agreement with human domain raters (error bars: 95\% task-level bootstrap CIs). The Agentic Autorater reaches 96.2\% agreement ($A^{\mathrm{HA}}$), matching 96.4\% of the 99.8\% inter-human ceiling ($A^{\mathrm{HH}}$).}
  \label{fig:human-agreement}
\end{figure}

\paragraph{Results.}
The Agentic Autorater reaches $A^{\mathrm{HA}} = 96.2\%$ $[94.7, 97.6]$ alongside an inter-human agreement ceiling of $A^{\mathrm{HH}} = 99.8\%$ $[99.6, 100.0]$---matching 96.4\% of human--human agreement (Figure~\ref{fig:human-agreement}; 99.7\% of criteria are unanimous across all three raters). Accounting for the 69.1\%/69.9\% weighted pass rates (expected chance agreement $57.6\%$), weighted pairwise Cohen's $\kappa$ is $0.911$ $[0.873, 0.943]$ ($0.919$ $[0.882, 0.950]$ on majority verdicts; inter-human Fleiss' $\kappa = 0.996$). Such near-unanimous expert agreement is expected because MKA rubrics specify objective, factual checks (computed values, tolerances, required fields, and explicit formatting rules). At the task level, autorater weighted rubric scores track human scores closely, with Pearson $r = 0.966$ and Spearman $\rho = 0.931$.

\paragraph{Error breakdown.}
Treating the human majority verdict as ground truth yields an autorater error rate of 3.8\% $[2.4, 5.4]$ per criterion (about one error in 26 criteria). Table~\ref{tab:autorater-errors} decomposes this 3.8\% error rate into \emph{false passes} (the deliverable is wrong or missing a required item, yet the autorater marks the criterion passed; 2.4\%) and \emph{false fails} (the deliverable is correct, yet the autorater marks the criterion failed; 1.4\%). False passes mainly reflect boundary leniency, such as crediting alternative or non-primary values when the model's committed answer differs, or accepting semantically equivalent content when a required label is omitted. Conversely, false fails stem from overly strict string or structural matching where domain experts accept equivalent formulations, or from static code inspection of interactive deliverables without live execution.

\begin{table}[t]
  \centering
  \small
  \caption{Decomposition of the 3.8\% per-criterion autorater error rate against the human majority verdict (95\% task-level bootstrap CIs). Conditional rates normalize by the human-failed (34.2\%) and human-passed (65.8\%) base rates.}
  \label{tab:autorater-errors}
  \begin{tabular}{llcc}
    \toprule
    \textbf{Error Type} & \textbf{Definition (vs.\ Human Majority)} & \textbf{\% of All Criteria} & \textbf{Conditional Rate (\%)} \\
    \midrule
    False pass & Human fails, autorater passes & 2.4 $[1.3, 3.8]$ & 7.1 $[3.6, 11.4]$ \\
    False fail & Human passes, autorater fails & 1.4 $[0.8, 2.0]$ & 2.1 $[1.2, 3.1]$ \\
    \midrule
    \textbf{Total error} & Any disagreement with human majority & \textbf{3.8} $[2.4, 5.4]$ & --- \\
    \bottomrule
  \end{tabular}
\end{table}

\subsection{Quality and Efficiency Metrics}
\label{sec:metrics}

We evaluate each model on deliverable quality and token efficiency across all 200 tasks.
All metrics are task-level means with tasks weighted equally. Rollouts that fail due to transient infrastructure or API errors are re-run. Sessions that hit the turn or wall-clock limit are not re-run and are graded on any outputs produced before termination; rollouts that produce no output (such as safety refusals) score zero on all criteria.

\paragraph{Weighted Satisfaction Rate (WSR).}
Each task $i$ has rubric criteria $j$ with expert weights $w_{ij} \in \{1, \dots, 5\}$
and binary autorater verdicts $s_{ij} \in \{0, 1\}$. Task WSR is the weighted share of
satisfied criteria:
\begin{equation}
  \mathrm{WSR}_i = \frac{\sum_j w_{ij}\, s_{ij}}{\sum_j w_{ij}}.
  \label{eq:task-wsr}
\end{equation}
WSR scores each criterion directly from $w_{ij}$ and $s_{ij}$ without gating on the
primary/secondary labels or prerequisite DAGs from Section~\ref{sec:dataset}; those
annotations structure rubric authoring and quality control, and we retain them in the
released metadata for diagnostic analysis and optional gated scoring. We do not report task-level all-pass rate
because it is brittle and has high variance: with 70+ criteria per task, a single missed
minor check or autorater flip zeroes out downstream criteria or the entire task.

\paragraph{Climb and regression splits.}
Domain experts author rubrics covering every task requirement, including basic
formatting checks that current models pass routinely. Using a fixed panel of four
prior-generation models (Gemini~3.8 Flash, Gemini~3.6 Flash, Claude Sonnet~5.5 and
GPT-6.1 Sol), we label a criterion \emph{regression} if every model in the panel passes
it in at least 2 of 3 runs (6,396 criteria across 180 tasks), and \emph{climb} otherwise
(7,888 criteria across all 200 tasks). Tasks that consistently run out of context for the
model panel are labeled all-\emph{climb}. Because this split is derived from autorater
verdicts, a small fraction of criteria the panel satisfies may also be labeled \emph{climb}
due to judge false fails (Section~\ref{sec:human-correlation}); the 2-of-3 rule dampens
single-run flips, and the frozen split applies equally to all evaluated models. Computing
Eq.~\eqref{eq:task-wsr} over each subset yields three task-level scores: \textbf{Climb-WSR}
(headline; frontier capability), \textbf{Regression-WSR} (retention of skills that
prior-generation models already have, averaged over the 180 tasks with regression criteria)
and \textbf{Overall WSR} (overall deliverable quality).

\section{Experiments and Results}
\label{sec:results}
We evaluate Gemini~4 Carbon alongside leading external frontier models: Claude Fable~5.2,
Claude Opus~5.5 and GPT-6 Astra. Each model runs once per task in the same Stirrup harness,
container environment and turn budget (Section~\ref{sec:harness}) and is graded in a single
pass by the Agentic Autorater (Section~\ref{sec:autorater}). None of the evaluated models is part of the
panel used to define the climb and regression splits (Section~\ref{sec:metrics}).

\subsection{Model Quality Comparison}
\label{sec:quality-results}
\begin{figure}[t]
  \centering
  \includegraphics[width=0.96\textwidth]{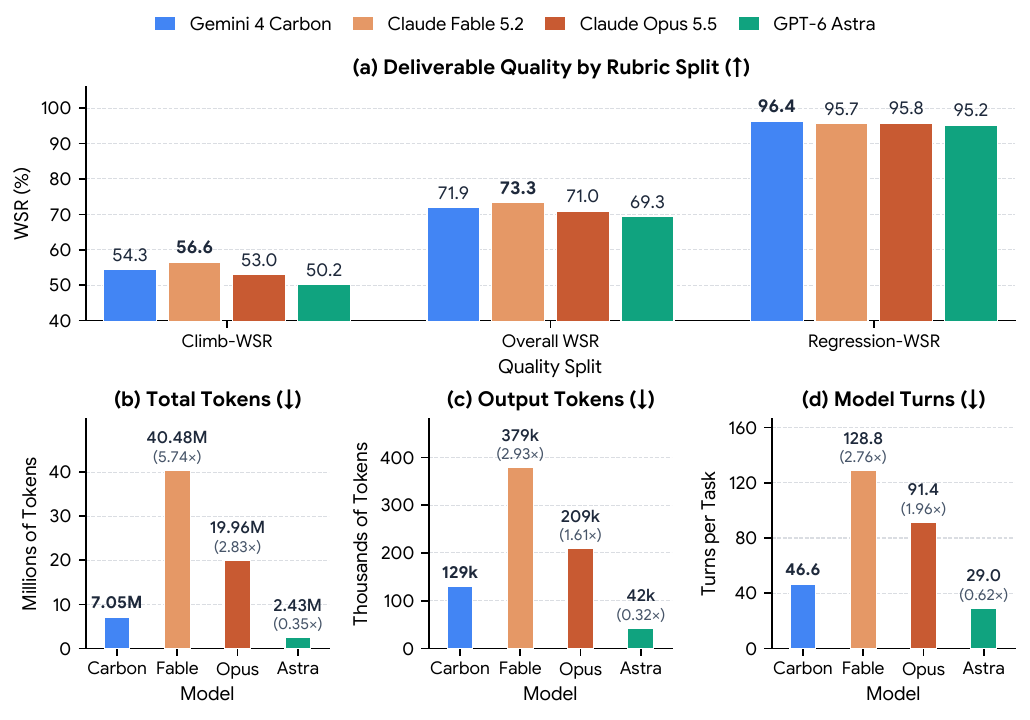}
  \caption{Deliverable quality across rubric splits (\textbf{a}; task-level means) and mean
  total tokens, output tokens and model turns per task (\textbf{b}--\textbf{d}; multipliers
  relative to Gemini~4 Carbon).}
  \label{fig:quality_efficiency}
\end{figure}

Table~\ref{tab:quality-results} and Figure~\ref{fig:quality_efficiency}a compare
task-level Climb-WSR, Overall WSR and Regression-WSR. All four models saturate
Regression-WSR above 95\% (95.17\%--96.36\%), confirming that frontier differences are
concentrated in Climb-WSR. Claude Fable~5.2 (56.56\% Climb-WSR, 73.26\% Overall WSR) and
Gemini~4 Carbon (54.34\% Climb-WSR, 71.89\% Overall WSR) achieve the highest deliverable
quality. Under paired task-level bootstrap, the top three models---Claude Fable~5.2,
Gemini~4 Carbon, and Claude Opus~5.5 (53.01\% Climb-WSR, 70.98\% Overall WSR)---are
statistically indistinguishable on Climb-WSR (Fable~5.2 vs.\ Carbon: $+2.22$~pp,
95\% CI $[-0.94, +5.49]$; Carbon vs.\ Opus~5.5: $+1.33$~pp, 95\% CI
$[-1.70, +4.41]$), whereas both Fable~5.2 and Gemini~4 Carbon significantly outperform
GPT-6 Astra (50.24\% Climb-WSR, 69.34\% Overall WSR; Carbon vs.\ Astra: $+4.10$~pp
Climb-WSR, 95\% CI $[+1.00, +7.11]$). Even for the strongest frontier models, Climb-WSR
remains around 54\%--57\%, leaving substantial headroom to evaluate multimodal agent
progress.

\begin{table}[!htbp]
  \centering
  \small
  \caption{Deliverable quality on MKA (task-level means $\pm$ 95\% CI, \%).}
  \label{tab:quality-results}
  \begin{tabular}{lccc}
    \toprule
    \textbf{Model} & \textbf{Climb-WSR} & \textbf{Overall WSR} & \textbf{Regression-WSR} \\
    \midrule
    Claude Fable 5.2 & \textbf{56.56} {\scriptsize$\pm$3.63} & \textbf{73.26} {\scriptsize$\pm$2.45} & 95.71 {\scriptsize$\pm$1.68} \\
    Gemini 4 Carbon  & 54.34 {\scriptsize$\pm$3.24} & 71.89 {\scriptsize$\pm$2.15} & \textbf{96.36} {\scriptsize$\pm$0.91} \\
    Claude Opus 5.5  & 53.01 {\scriptsize$\pm$3.54} & 70.98 {\scriptsize$\pm$2.46} & 95.81 {\scriptsize$\pm$1.32} \\
    GPT-6 Astra      & 50.24 {\scriptsize$\pm$3.25} & 69.34 {\scriptsize$\pm$2.10} & 95.17 {\scriptsize$\pm$1.07} \\
    \bottomrule
  \end{tabular}

  \vspace{12pt}

  \caption{Mean per-task token usage, agent turns, runtime and list-price API cost (with prompt caching) on MKA. Output tokens sum reasoning and answer tokens for every model.}
  \label{tab:efficiency-results}
  \begin{tabular}{lcccccc}
    \toprule
    \textbf{Model} & \textbf{Total Tok.} & \textbf{Input Tok.} & \textbf{Output Tok.} & \textbf{Turns} & \textbf{Time (min)} & \textbf{API Cost} \\
    \midrule
    Claude Fable 5.2 & 40.48M & 40.10M & 378.9k & 128.8 & 83.2 & \$35.28 \\
    Gemini 4 Carbon  & 7.05M  & 6.92M  & 129.5k & 46.6  & 29.2 & \textbf{\$5.63} \\
    Claude Opus 5.5  & 19.96M & 19.75M & 208.8k & 91.4  & 33.8 & \$10.29 \\
    GPT-6 Astra      & \textbf{2.43M} & \textbf{2.39M} & \textbf{42.0k} & \textbf{29.0} & \textbf{18.5} & \$6.03 \\
    \bottomrule
  \end{tabular}
\end{table}

\subsection{Model Efficiency Comparison}
\label{sec:efficiency-results}
Table~\ref{tab:efficiency-results} and Figures~\ref{fig:quality_efficiency}b--d compare
mean per-task token usage, trajectory length, runtime and estimated API cost at list
prices (accounting for each provider's prompt-cache read/write pricing, where 93.7\%--97.7\%
of cumulative input tokens are served from cache; output tokens include both reasoning and
visible answer tokens for all models). Trajectory efficiency varies by over $16\times$
across frontier models:
\begin{itemize}
  \item \textbf{Claude Fable~5.2} consumes $5.74\times$ as many total tokens as Gemini~4
        Carbon (40.48M vs.\ 7.05M), $2.93\times$ as many output tokens (378.9k vs.\ 129.5k,
        with 286.8k vs.\ 76.6k reasoning tokens), $2.76\times$ as many turns (128.8
        vs.\ 46.6), and $2.85\times$ the wall-clock time (83.2 vs.\ 29.2 minutes per task).
        At provider list prices, this corresponds to a $6.26\times$ higher API cost (\$35.28
        vs.\ \$5.63 per task) for a statistically non-significant $+2.22$~pp Climb-WSR
        difference.
  \item \textbf{Claude Opus~5.5} likewise uses $2.83\times$ as many total tokens (19.96M),
        $1.61\times$ as many output tokens (208.8k over 91.4 turns) and $1.83\times$ higher
        API cost (\$10.29 per task) than Gemini~4 Carbon while scoring $1.33$~pp lower on
        Climb-WSR.
  \item \textbf{GPT-6 Astra} uses the fewest tokens and turns ($0.35\times$ Carbon's total
        tokens: 2.43M total and 42.0k output over 29.0 turns in 18.5 minutes), yet because
        of higher per-token list pricing it still costs slightly more per task than Gemini~4
        Carbon (\$6.03 vs.\ \$5.63) while scoring 4.10~pp lower on Climb-WSR.
\end{itemize}

\begin{figure}[!htbp]
  \centering
  \includegraphics[width=0.55\textwidth]{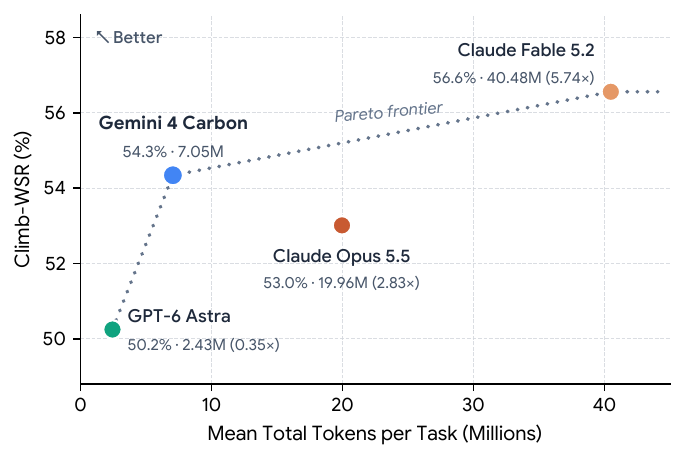}
  \caption{Deliverable quality (Climb-WSR) versus mean total tokens per task across MKA tasks (upper-left is better). GPT-6 Astra, Gemini~4 Carbon and Claude Fable~5.2 trace the token Pareto frontier, with Gemini~4 Carbon using fewer tokens than Claude Opus~5.5 at similar quality, and sitting at the knee before Fable~5.2's $5.74\times$ token jump.}
  \label{fig:pareto_climb_tokens}
\end{figure}

Figure~\ref{fig:pareto_climb_tokens} plots Climb-WSR against total token usage: GPT-6
Astra, Gemini~4 Carbon and Claude Fable~5.2 form the token Pareto frontier, with Gemini~4
Carbon offering the best quality-per-token among the top scoring models, and achieving the lowest API cost per task (\$5.63).

\subsection{Role of Visual Perception}
\label{sec:vision-ablation}
To test how much MKA relies on genuine visual perception rather than text extraction, we evaluate Gemini~4 Carbon with and without the \texttt{view\_image} tool under the identical 250-turn Stirrup setup, and run the same ablation on GDPval~\citep{patwardhan2025gdpval}. Without \texttt{view\_image}, the agent retains \texttt{code\_exec}, \texttt{web\_search}, and all pre-installed parsing, OCR, and computer-vision libraries, so it can still extract embedded document text, parse spreadsheets, and write custom image-analysis scripts. On GDPval, we grade both arms with the same Agentic Autorater (Section~\ref{sec:autorater}) against the expert-authored rubrics released with the public GDPval gold subset and report Overall WSR ($88.1\%$ w/ vs.\ $88.2\%$ w/o \texttt{view\_image}).

\paragraph{MKA vs.\ GDPval.}
Table~\ref{tab:vision-ablation} shows a sharp split between the two benchmarks. On GDPval, removing \texttt{view\_image} changes neither the deliverable's analytical quality ($+0.08$~pp Overall WSR, $95\%$~CI $[-0.62, +0.78]$) nor compute ($0.95\times$ turns, $1.05\times$ output tokens, and $0.89\times$ input tokens), because task-critical inputs reside in spreadsheets and machine-readable text. On MKA, removing \texttt{view\_image} reduces Overall WSR by $4.58$~pp ($[-5.51, -3.65]$) and Climb-WSR by $5.94$~pp ($[-7.39, -4.49]$)---more than $3\times$ the drop on Regression-WSR ($-1.96$~pp)---while inflating trajectories to $1.28\times$ model turns, $1.40\times$ output tokens, and $1.85\times$ input tokens.

\begin{table}[t]
  \centering
  \small
  \setlength{\tabcolsep}{6pt}
  \caption{Tool ablation (w/ vs.\ w/o \texttt{view\_image}) for Gemini~4 Carbon on MKA and GDPval. Removing \texttt{view\_image} leaves GDPval unchanged in both quality and cost, whereas MKA suffers a significant Climb-WSR drop alongside steep token and turn inflation.}
  \label{tab:vision-ablation}
  \begin{tabular}{lcccccc}
    \toprule
     & \multicolumn{3}{c}{\textbf{Quality $\Delta$ (w/o $-$ w/ \texttt{view\_image}, pp)}} & \multicolumn{3}{c}{\textbf{Compute Ratio (w/o $/$ w/ \texttt{view\_image})}} \\
    \cmidrule(lr){2-4} \cmidrule(lr){5-7}
    \textbf{Benchmark} & \textbf{Climb-WSR} & \textbf{Overall WSR} & \textbf{Reg.-WSR} & \textbf{Input Tok.} & \textbf{Output Tok.} & \textbf{Turns} \\
    \midrule
    MKA    & $\mathbf{-5.94}$ & $\mathbf{-4.58}$ & $\mathbf{-1.96}$ & $\mathbf{1.85\times}$ & $\mathbf{1.40\times}$ & $\mathbf{1.28\times}$ \\
    GDPval & ---              & $+0.08$ (ns)     & ---              & $0.89\times$          & $1.05\times$          & $0.95\times$ \\
    \bottomrule
  \end{tabular}
\end{table}

\paragraph{Why strong coding can compensate---but cannot replace---vision.}
Although a $-5.94$~pp Climb-WSR drop is substantial and statistically significant, it is smaller than one might expect on a visually grounded benchmark. Inspecting trajectories without \texttt{view\_image} reveals that Gemini~4 Carbon partially compensates through pure persistence using code to create workarounds. Surprisingly, it goes far beyond standard text extraction and OCR, and starts rendering image crops as ASCII art or unicode Braille blocks in its terminal, or falls back to complicated and bespoke classical computer vision pipelines with OpenCV. Two case studies (Appendix~\ref{sec:case-studies}) illustrate the resulting split:
\begin{itemize}
  \item \textbf{Partial recovery at steep cost:} On clean schematics (e.g., mechanical CAD drawings), ASCII rendering and OpenCV measurements recover primary geometry in some runs, but miss fine clearance hazards and require $1.8\times$--$4.6\times$ more tokens and turns.
  \item \textbf{Outright perceptual failure:} On natural photographs and medical imaging (e.g., road-inspection photos, clinical ultrasounds and radiographs, or circuit-board photos), dozens of OpenCV and ASCII-art scripts still misread subtle visual structure, dropping Climb-WSR by $34$--$72$~pp despite $1.8\times$--$5.8\times$ higher compute.
\end{itemize}

\section{Related Work}
\label{sec:related_work}
MKA builds on a long tradition of benchmarks that have driven progress across two largely separate tracks: static multimodal perception over documents, charts, and diagrams, and autonomous tool use in coding, web, and office environments. Below, we situate MKA alongside prior work across workplace deliverables, spreadsheet workflows, multimodal question answering, active visual reasoning, agentic search, and computer-use grounding (summarized in Table~\ref{tab:benchmark_comparison}).
\paragraph{Workplace Deliverables.}
Building on repository-level coding and CLI benchmarks~\citep{jimenez2024swebench,yang2025swebenchmultimodal,merrill2026terminalbench}, multi-application office and simulated company environments~\citep{wang2024officebench,xu2024theagentcompany}, and domain-expert rubric evaluations~\citep{starace2025paperbench,arora2025healthbench,vidgen2025apex}, recent suites evaluate agents on realistic project deliverables. APEX-Agents~\citep{vidgen2026apexagents} places agents in multi-file worlds across banking, consulting, and law, grading outputs via per-criterion LLM-judge calls (\texttt{Pass@1}). GDPval~\citep{patwardhan2025gdpval} spans 44 occupations and grades deliverables primarily via blinded pairwise preference. Although its workspaces include PDFs and slide decks, task-critical signal can largely be extracted via Python scripts and OCR without visual inspection.
\paragraph{Tabular and Spreadsheet Workflows.}
Dedicated spreadsheet benchmarks isolate multi-step code execution over complex workbooks without requiring visual perception or non-tabular deliverables. GDP.xlsx~\citep{surgeai2026gdpxlsx} equips agents with sandboxed Python tools over a single workbook to produce a text answer, while SpreadsheetBench~2~\citep{zhu2026spreadsheetbench2,ma2024spreadsheetbench} evaluates \texttt{.xlsx} modeling, debugging, and chart creation without cross-modal inputs.
\paragraph{Multimodal and Document Question Answering.}
Foundational document and chart QA benchmarks~\citep{mathew2021docvqa,masry2022chartqa,ma2024mmlongbenchdoc} and harder frontier suites such as MMMU~\citep{yue2024mmmu}, CharXiv~\citep{wang2024charxiv}, and MMMU-Pro~\citep{yue2024mmmupro} require strong visual perception, but present pre-selected images or pages in a single, tool-free turn and grade multiple-choice or short text answers. Diagnostic suites including MMStar~\citep{chen2024mmstar}, BLINK~\citep{fu2024blink}, VLMs Are Blind~\citep{rahmanzadehgervi2024vlmsareblind}, and VLMs Are Biased~\citep{vo2025vlmsarebiased} further show that VLMs often bypass pixels via linguistic priors or fail at basic 2D geometric tracing. Extending document QA to enterprise files, Surge AI's GDP.pdf~\citep{garre2026gdppdf} evaluates atomic criterion rubrics over dense, multi-document professional PDFs, yet still pre-attaches source PDFs in a single-turn prompt and grades a text response via a text-only judge---omitting iterative tool use, cross-format synthesis, and binary deliverable verification.
\paragraph{Frontier VLMs and Active Visual Tool Use.}
Frontier multimodal models across proprietary~\citep{deepmind2025gemini3,anthropic2026claudeopus46,openai2026gpt5,meta2026muse} and open-weight~\citep{bai2025qwen3vl,zhu2025internvl3,meta2025llama4} families pair native visual perception with code-driven agent scaffolds~\citep{wang2024codeact}, and recent models across labs---from Google's Agentic Vision~\citep{google2026agenticvision} to Claude~\citep{anthropic2026claudeopus46}, Meta Muse~\citep{meta2026muse}, Qwen~\citep{bai2025qwen3vl}, and OpenAI~o3~\citep{openai2025o3}---actively inspect visual inputs by cropping, zooming, and transforming images inside chain-of-thought. MKA tests whether these active perceptual capabilities transfer to multi-file enterprise workspaces where agents must choose when to write code versus when to inspect pixels.
\paragraph{Agentic Search and Retrieval.}
Following general tool-use assistants such as GAIA~\citep{mialon2024gaia}, deep-research benchmarks evaluate multi-step web search over text (BrowseComp;~\citealt{wei2025browsecomp}) and visual web content (MM-BrowseComp;~\citealt{li2025mmbrowsecomp}). By design, however, these benchmarks constrain outputs to a single short string or number to permit exact-match grading, leaving open-ended multi-file synthesis untested.
\paragraph{GUI and Computer-Use Environments.}
Browser and desktop benchmarks~\citep{xie2024osworld,li2025screenspotpro} evaluate agents acting through low-level pixel coordinates (\texttt{mouse\_move}, \texttt{click(x, y)}) or DOM actions, verifying completion via deterministic post-hoc scripts that check exact file diffs, DOM states, or database records. While effective for closed-form UI tasks, model failures are dominated by brittle interface grounding, and deterministic state scripts cannot evaluate open-ended reports, executive slide decks, or valid layout trade-offs. Building on LLM-as-a-judge~\citep{zheng2023mtbench}, MLLM-as-a-Judge~\citep{chen2024mllmasajudge}, and Agent-as-a-Judge~\citep{zhuge2025agentasajudge} paradigms alongside classical inter-rater agreement~\citep{cohen1960coefficient,fleiss1971measuring}, MKA decouples domain-level multimodal reasoning from GUI clicking while grading binary deliverables with a sandboxed visual and programmatic autorater.

\section{Limitations}
\label{sec:limitations}
Each model runs once per task and is graded in a single autorater pass, so our confidence
intervals capture task sampling but not run-to-run agent or grader variance. The human agreement
study covers 50 tasks and focuses on high-weight criteria (weights 3--5) with a balanced pass/fail
mix, so the 96.2\% agreement may not transfer exactly to low-weight formatting checks. The vision
ablation is run on Gemini~4 Carbon only; other models may rely on vision to different degrees. The
sandbox has internet access, which mirrors real deployments but may let agents retrieve some
information online instead of extracting it visually. The climb/regression panel includes two
earlier Gemini models, so the split reflects what that panel already solves. Finally, API costs
use provider list prices at the time of writing and will change.

\section{Conclusion}
\label{sec:conclusion}
Real-world knowledge work is inherently visual, requiring professionals to reason over blueprints, schematics, maps, charts, and scanned documents while verifying the layout of the files they produce. MKA brings this reality into agent evaluation across 200 multi-file workflows and 10 sectors, paired with an Agentic Autorater that matches human domain experts on 96.2\% of criterion judgments in our expert study ($\kappa = 0.91$). Our findings show that native visual perception drives not only deliverable quality, but also compute efficiency: removing image-viewing tools degrades Climb-WSR while inflating turns ($1.28\times$) and output tokens ($1.40\times$) as agents fall back on brittle scripting workarounds, and frontier models exhibit an over $16\times$ spread in total token usage across multimodal trajectories (from 2.43M to 40.48M tokens per task), with Gemini~4 Carbon matching top deliverable quality at $5.74\times$ lower token usage ($6.26\times$ lower API cost) than Claude Fable~5.2 and $2.83\times$ lower token usage than Claude Opus~5.5. Yet while evaluated models exceed 95\% on routine regression checks, top Climb-WSR reaches just 56.6\%---leaving substantial headroom for agents that natively coordinate visual inspection, code execution, and visual self-verification.

\clearpage
\section*{Acknowledgments}
\noindent We thank our colleagues across Gemini and the broader Google DeepMind team for their foundational research, infrastructure, and helpful feedback throughout this project. We are especially grateful to the following individuals and partners for their contributions:
\begin{itemize}
  \item \textbf{External Evaluation \& Domain Experts:} We thank Artificial Analysis for partnering on standardized harness evaluations (Stirrup) and hosting the public MKA leaderboard, and we thank the domain specialists across all ten verticals whose expertise made this benchmark possible.
  \item \textbf{Initial Autorater Implementation:} Balaji Lakshminarayanan, Xi Liu, and Ben Withbroe.
  \item \textbf{Core Serving \& Infrastructure Support:} Steven Guh, Calin Juravle, Nam Luu Nhat, and Sehoon Shon.
  \item \textbf{Program Management Support:} Slavica Anda\v{c}i\'c, Sumit Bagri, Shruti Garg, Tiffany Loer, Kate Olszewska, and Nandita Sethi.
  \item \textbf{AI Co-Scientist \& Editorial Support:} We thank Gemini Argon and Gemini Carbon for serving as AI co-authors, coding collaborators, and editors during the design of the benchmark, analysis of results, and writing of this manuscript.
\end{itemize}

\clearpage
\bibliography{main}
\clearpage
\section*{Appendix}
\addcontentsline{toc}{section}{Appendix}
\appendix
\section{Agentic Autorater Prompt}
\label{sec:autorater-prompt}

Listing~\ref{lst:autorater-prompt} shows the verbatim prompt template used by the Agentic Autorater (Section~\ref{sec:autorater}). At runtime, the grading harness populates five placeholders:
\begin{itemize}
  \item \texttt{\{deliverables\_line\}}: Comma-separated list of harvested deliverable filenames staged in \texttt{/workspace} (or \texttt{(none - the agent submitted no files)}).
  \item \texttt{\{chat\_line\}}: Path to the staged transcript of the candidate's final text response when non-empty (or \texttt{(none - the agent said nothing in its final message)}).
  \item \texttt{\{task\_prompt\_section\}}: The verbatim task prompt given to the candidate agent, prefixed by \texttt{TASK PROMPT GIVEN TO THE AGENT:}.
  \item \texttt{\{rubrics\_content\}}: The task's rubric JSON mapping each \texttt{"criterion N"} key to its \texttt{description} and integer \texttt{weight}.
  \item \texttt{\{output\_warning\_section\}}: Empty when the candidate submitted at least one deliverable file, or Listing~\ref{lst:autorater-chat-only} when the candidate responded only in chat.
\end{itemize}
Calling \texttt{submit\_verdicts} validates that every rubric criterion has a structured verdict and terminates the grading session.

\begin{lstlisting}[
  caption={Agentic Autorater prompt template.},
  label={lst:autorater-prompt},
  basicstyle=\ttfamily\footnotesize,
  breaklines=true,
  breakatwhitespace=true,
  breakindent=0pt,
  columns=fullflexible,
  keepspaces=true,
  frame=single,
  xleftmargin=4pt,
  xrightmargin=4pt
]
You are grading an agent's work on a task. You are given two kinds of evidence:
the files the agent submitted, and the agent's final chat message. Your task is to verify whether the agent's work satisfies the criteria
provided below.

The files have been written to the directory /workspace in your environment.
Use the code_exec tool for shell commands (Python scripts, LibreOffice,
pdftoppm, ...) and the view_image tool to look at images or rendered pages.
To inspect a PDF or Office file visually, render its pages to PNG first (e.g.
pdftoppm -png -r 150 <file>.pdf <prefix>) and view the PNGs.
IMPORTANT: When viewing several images, view them one at a time, never output
multiple view_image calls in a single response.

STRATEGY GUIDE FOR OFFICE DOCUMENTS:
When an output artifact is a Microsoft Office file (.docx, .pptx, .xlsx), choose
the most effective evaluation strategy (or a combination) based on the specific
rubric criteria. In most cases, START by rendering the document (Option A): the
extraction libraries in Option B routinely fail to "see" content that a human
grader sees (computed cell values, theme colors, text inside tables), which is a
common cause of FALSE NEGATIVES.
Option A (Rendered Evaluation - PREFER THIS for content presence, values,
  colors, and visual layout): Convert the document to PDF, render its pages to PNG,
  and inspect them with the view_image tool. The rendered PDF shows computed cell values, theme/fill colors,
  text inside tables, charts, and the visual layout that programmatic extraction
  often misses. Use the code_exec tool to run LibreOffice:
    soffice --headless --norestore --nolockcheck --convert-to pdf <filename> --outdir /workspace
  Then render the PDF to PNG (pdftoppm -png -r 150 <file>.pdf <prefix>) and use
  the view_image tool on the pages.
  For SPREADSHEETS (.xlsx), ALSO export each sheet to CSV, because a wide or long
  table can overflow the PDF page and be silently clipped, and CSV yields exact,
  unrounded values:
    soffice --headless --norestore --nolockcheck --convert-to csv <filename> --outdir /workspace
  Caveat for computed values: LibreOffice does not always recalculate formulas
  when converting a programmatically-generated .xlsx that has no cached results,
  so a computed total may render blank. If an expected value is blank, do NOT
  conclude it is missing - also read the formula (openpyxl data_only=False) or
  compute it yourself before deciding.
  For PAGE COUNT criteria, see the PAGE COUNT section below for the correct method per format.
Option B (Programmatic Evaluation - for metadata, styles, and exact extraction):
  If the criteria require strict metadata verification, style checks, semantic
  structural validation, or exact text/number extraction (e.g., "does it use the
  H1 style", exact formula text, "read massive tables"), write Python scripts
  (e.g., python-docx, python-pptx, openpyxl) or parse the ZIP/XML directly. Watch
  for these library pitfalls that commonly cause FALSE NEGATIVES; never declare a
  value or section "not present" because of them:
  * Formula cells (openpyxl): load_workbook(data_only=True) returns None for a
    formula cell that has no cached result (common in files generated
    programmatically rather than saved by Excel). If totals, aggregates, or summary
    values read as None, also read the file with data_only=False to see the
    formula, and/or render the file (Option A) to obtain the computed values,
    BEFORE concluding the value is missing.
  * Cell colors (openpyxl): theme-based fills are NOT exposed by
    cell.fill.start_color.index (it reports '00000000' or a theme index, not an
    RGB). Do not infer "no color" from that check alone; inspect fill.patternType
    and the theme/tint attributes, or evaluate the color visually via Option A.
  * Table/box text (python-docx): doc.paragraphs does NOT include text inside
    tables, headers, footers, or text boxes. Before concluding a value or answer
    is "not stated", also iterate doc.tables (and each cell's .paragraphs) and
    check headers/footers - answers in a table cell are otherwise easily missed.

EVIDENCE AVAILABLE TO YOU

1. Deliverables submitted by the agent (files the agent chose to produce):
{deliverables_line}

2. The agent's final chat message. This is a TRANSCRIPT OF WHAT THE AGENT SAID,
   NOT a file the agent created. It has been written to a file only so that you
   can read it with your tools. Never treat its existence as evidence that the
   agent produced a document, and never list it as a deliverable:
{chat_line}{task_prompt_section}

Here is the rubrics JSON containing the criteria to verify:

{rubrics_content}

EVALUATION OF DELIVERABLES AND CHAT RESPONSE:
Always include any artifact submitted by the agent in your
evaluation. Based on the task prompt given to the agent, assess whether it is
appropriate to grade the submitted artifact along with the final chat response:
If the task prompt asked for, implied, or is naturally satisfied by creating a
  deliverable (such as a report, policy brief, memo, review, spreadsheet,
  presentation, code file, plain-text analysis, summary document, etc.), grade
  the submitted artifact(s) and chat response together. Substantive content
  delivered in an artifact fully satisfies criteria requiring that information,
  even if the chat response is concise or acts as a transmittal summary.
Do NOT penalize the agent for delivering content in a submitted file rather
  than directly in chat unless the task prompt explicitly and strictly forbade
  creating or attaching files.
If the task prompt explicitly and strictly prohibited creating or attaching
  files (e.g., "Do not create any files", "Output only in console without creating
  files"), submitting a file violates that instruction.
If no files were submitted, grade the criteria against the chat response on
  its merits.

GENERAL GRADING PRINCIPLES (apply to EVERY criterion):
1. Honor stated tolerances. When a criterion gives an acceptable range or an
   approximate value (e.g. "acceptable range X to Y", "approximately N", "+/- P%"),
   mark it satisfied if the artifact's value falls within that range, even if it is
   not the exact target. Compute and compare the value with a tool rather than
   relying on visual estimation.
2. Before declaring content ABSENT, inspect the whole artifact and rule out
   tool/library quirks (see the spreadsheet pitfalls above). Content may live in a
   later section, another sheet or tab, a merged cell, or a computed/summary cell.
3. PAGE COUNT -- use the correct method for each format:
   - PDF: page count is baked into the file format. Use:
       pdfinfo <file>.pdf | grep Pages
     or python3 -c "import fitz; print(fitz.open('<file>.pdf').page_count)"
   - PPTX: each slide is a separate file inside the ZIP archive. Count them:
       unzip -l <file>.pptx | grep -c 'ppt/slides/slide[0-9]'
     This is 100% reliable regardless of metadata.
   - DOCX: page count is NOT stored in the file -- it is computed at render time.
     Convert to PDF with LibreOffice and count the PDF pages as an ESTIMATE:
       soffice --headless --norestore --nolockcheck --convert-to pdf <file>.docx --outdir /workspace
       pdfinfo <file>.pdf | grep Pages
     Note: LibreOffice may render slightly more pages than Microsoft Word for the
     same content (different font metrics and line spacing). If the PDF page count
     exceeds the criterion limit by exactly 1 page, treat it as a BORDERLINE PASS
     -- the extra page is likely a rendering difference, not a genuine violation.

{output_warning_section}

After all verifications complete, call the submit_verdicts tool once with exactly one
verdict per criterion. Each verdict has these fields:
"criterion": the criterion key from the rubrics JSON, e.g. "criterion 1"
"justification": string with your reasoning, written before you decide
"satisfied": boolean (true or false), decided from that reasoning
"sources": list of filenames you referenced to verify the criterion
\end{lstlisting}

\begin{lstlisting}[
  caption={Conditional \texttt{\{output\_warning\_section\}} block injected when the candidate submits no files.},
  label={lst:autorater-chat-only},
  basicstyle=\ttfamily\footnotesize,
  breaklines=true,
  breakatwhitespace=true,
  breakindent=0pt,
  columns=fullflexible,
  keepspaces=true,
  frame=single,
  xleftmargin=4pt,
  xrightmargin=4pt
]
IMPORTANT: The agent submitted NO files; it responded only in its final chat
message. Do NOT mark criteria unsatisfied merely because no file exists. Grade
each criterion against the chat message on its merits based on the task prompt:
if the task prompt strictly required a standalone document or file that was not
produced, criteria requiring that document cannot be satisfied; but substantive
content and answers delivered in the chat message must be evaluated fairly.
\end{lstlisting}

\clearpage
\section{Case Studies: Visual Perception vs.\ Programmatic Workarounds}
\label{sec:case-studies}

To illustrate how Gemini~4 Carbon adapts when \texttt{view\_image} is disabled---and where programmatic workarounds break down---we examine two representative tasks from the Stirrup vision ablation (Section~\ref{sec:vision-ablation}). Table~\ref{tab:case-studies} summarizes quality and efficiency across both tasks.

\begin{table}[!htbp]
  \centering
  \small
  \caption{Comparison of Gemini~4 Carbon w/ \texttt{view\_image} and w/o \texttt{view\_image} on two representative MKA tasks illustrating partial coding recovery versus perceptual collapse.}
  \label{tab:case-studies}
  \begin{tabular}{lccccc}
    \toprule
    \textbf{Case Study \& Arm} & \textbf{Climb-WSR} & \textbf{Overall WSR} & \textbf{Input Tok.} & \textbf{Output Tok.} & \textbf{Turns} \\
    \midrule
    1. CAD Review (w/ \texttt{view\_image}) & \textbf{76.3} & \textbf{88.7} & \textbf{1.36M} & \textbf{64.1k} & \textbf{21.7} \\
    1. CAD Review (w/o \texttt{view\_image}) & 63.3 & 82.4 & 6.21M & 113.9k & 51.0 \\
    \quad $\Delta$ / Ratio & $-13.0$ & $-6.2$ & $4.6\times$ & $1.8\times$ & $2.4\times$ \\
    \midrule
    2. Road Photo (w/ \texttt{view\_image}) & \textbf{86.7} & \textbf{96.0} & \textbf{4.87M} & \textbf{94.0k} & \textbf{43.0} \\
    2. Road Photo (w/o \texttt{view\_image}) & 14.3 & 64.7 & 18.06M & 207.5k & 76.5 \\
    \quad $\Delta$ / Ratio & $-72.4$ & $-31.3$ & $3.7\times$ & $2.2\times$ & $1.8\times$ \\
    \bottomrule
  \end{tabular}
\end{table}

\begin{figure}[!htbp]
  \centering
  \includegraphics[width=0.76\textwidth]{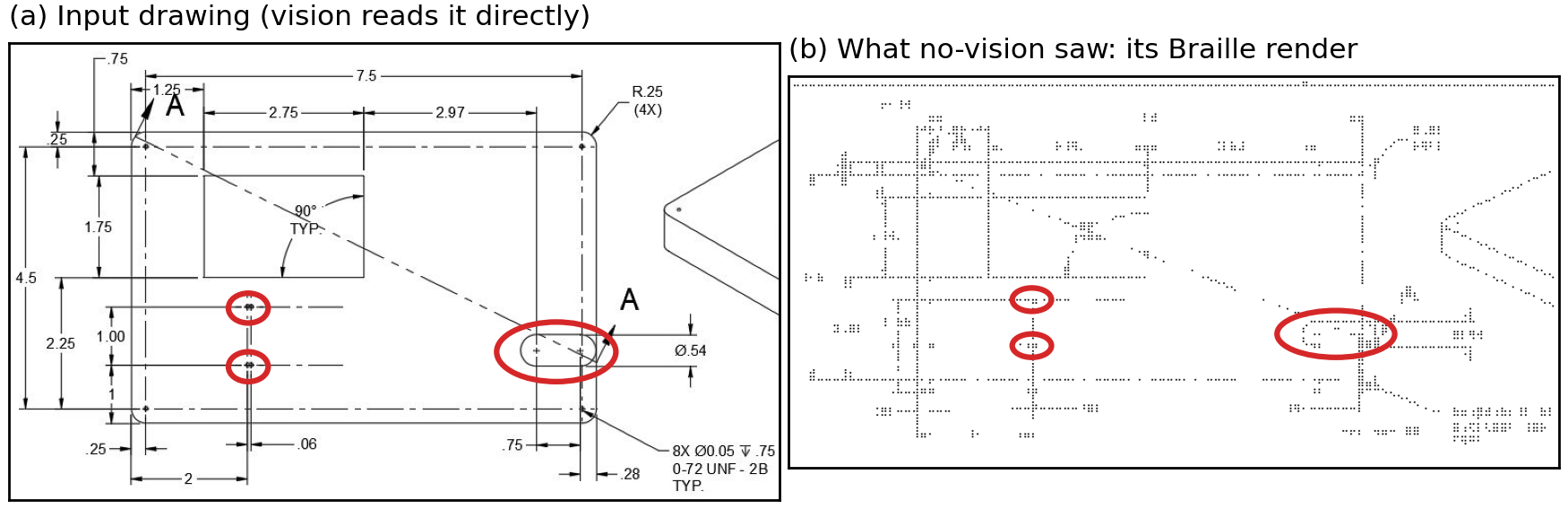}
  \caption{\textbf{Case Study 1: the workaround recovers most of the score at 2--5$\times$ the compute.}
  (a) The CAD drawing as the agent w/ \texttt{view\_image} sees it.
  (b) The Braille render that the agent w/o \texttt{view\_image} printed from \texttt{code\_exec}, cropped to the same region.
  Red rings mark the geometric features that the manufacturability judgment depends on.
  With \texttt{view\_image}, the agent catches both in 3/3 runs, while w/o \texttt{view\_image} catches them in 1/2.}
  \label{fig:case1_cad}
\end{figure}

\paragraph{Case Study 1: Coding workarounds recover most of the score at $2\times$--$5\times$ compute (Engineering).}
In this task, the agent evaluates whether a machined part shown only in a raster CAD drawing can be fabricated using a shop's equipment specifications, producing a structured manufacturability report (Figure~\ref{fig:case1_cad}).

\begin{itemize}
  \item \textbf{With \texttt{view\_image}:} Across 23 \texttt{view\_image} calls on targeted crops of the drawing, the agent directly reads part geometry, tolerances, and feature spacing, consistently flagging non-manufacturable design hazards across all three runs ($88.7\%$ Overall WSR, $76.3\%$ Climb-WSR).
  \item \textbf{Without \texttt{view\_image}:} Thirty of the agent's 49 \texttt{code\_exec} calls execute custom OpenCV pixel scans alongside ASCII and Unicode Braille renderings of cropped views. By measuring pixel spans between detected edges and pairing them with OCR'd dimension labels, the agent reconstructs an explicit pixel-to-inch scale factor and counts isometric features from Braille dot plots. This effort enables one run to tie the visual baseline's score ($86.2\%$ Overall WSR), while the second run misses tight feature-to-wall and hole-to-hole clearance violations ($78.7\%$ Overall WSR)---requiring $2.4\times$ the turns, $1.8\times$ the output tokens, and $4.6\times$ the cumulative input tokens.
\end{itemize}

\begin{figure}[!htbp]
  \centering
  \includegraphics[width=0.88\textwidth]{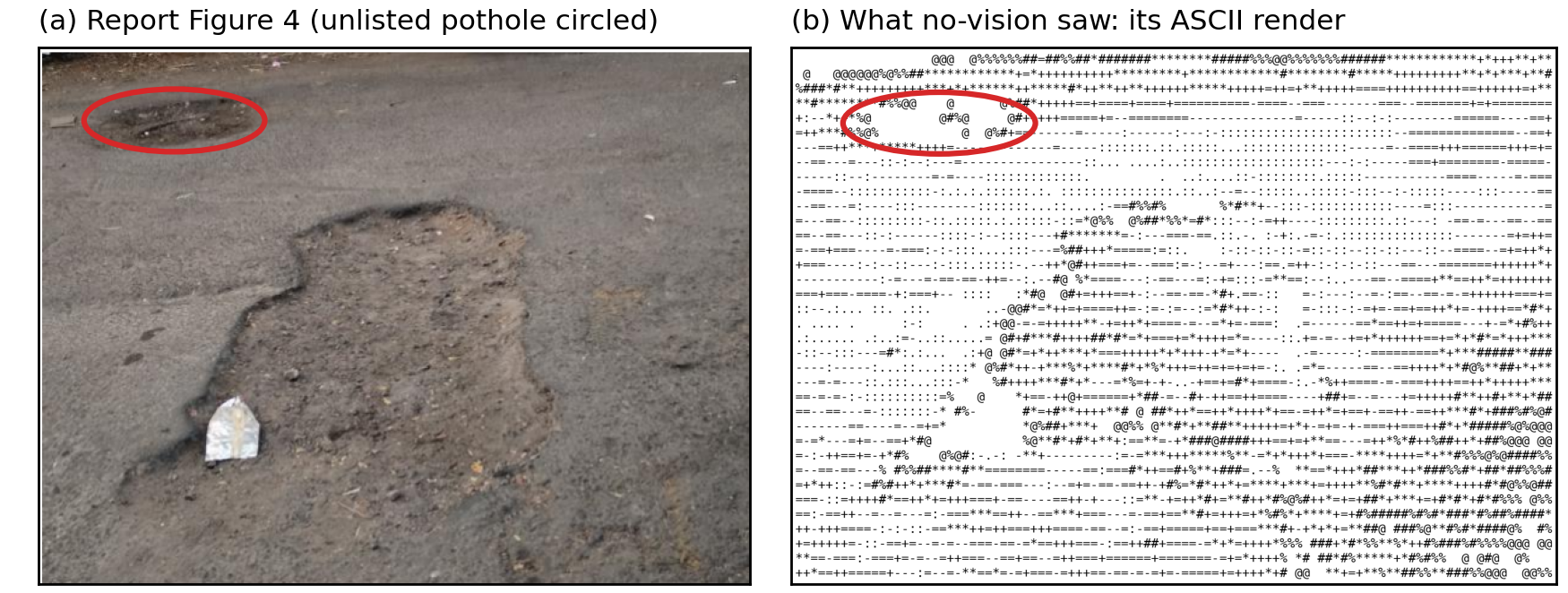}
  \caption{\textbf{Case Study 2: the workaround fails outright despite 2--4$\times$ the compute.}
  (a) A site photo from the inspection report. The red ring marks the region that holds the visual evidence the task asks for.
  (b) The ASCII luminance render that the agent w/o \texttt{view\_image} printed for the same photo.
  The evidence is present in the render as a dark cluster, but the agent does not recognize it.
  The agent w/ \texttt{view\_image} identifies it in every run. The agent w/o \texttt{view\_image} misses it in all runs, and Climb-WSR drops from 86.7 to 14.3.}
  \label{fig:case2_road}
\end{figure}

\paragraph{Case Study 2: Coding workarounds fail outright despite $2\times$--$4\times$ compute (Construction).}
In this task, the agent audits field photographs embedded in a road-inspection PDF against a visual distress classification manual and contractor rate schedules, producing a cost-estimate workbook that must catalog visible pavement defects omitted from the written log (Figure~\ref{fig:case2_road}).

\begin{itemize}
  \item \textbf{With \texttt{view\_image}:} Using 68 \texttt{view\_image} calls across extracted inspection photos and zoomed crops, the agent identifies undocumented surface depressions and matches their visual appearance to the corresponding pothole category in the reference manual ($96.0\%$ Overall WSR, $86.7\%$ Climb-WSR).
  \item \textbf{Without \texttt{view\_image}:} Sixty-seven of 84 tool calls run image-processing pipelines---extracting embedded photos, computing OpenCV morphological crack maps, rendering shaded ASCII art, running OCR, and matching ORB keypoints against manual figures. Unable to distinguish three-dimensional bowl depressions from surrounding asphalt texture in ASCII and edge maps, all runs conclude that the relevant figures contain no undocumented distresses and fail every criterion tied to the missing defects. Climb-WSR drops from $86.7\%$ to $14.3\%$ ($-72.4$~pp) despite $2.2\times$ the output tokens, $1.8\times$ the turns, and $3.7\times$ the input tokens.
\end{itemize}

\end{document}